\documentclass[conference]{IEEEtran}
\IEEEoverridecommandlockouts
\newcommand{\linebreakand}{%
  \end{@IEEEauthorhalign}
  \hfill\mbox{}\par
  \mbox{}\hfill\begin{@IEEEauthorhalign}
}
\usepackage[backend=bibtex]{biblatex}

\usepackage{amsmath,amssymb,amsfonts}
\usepackage{algorithmic}
\usepackage{graphicx}
\usepackage{textcomp}
\usepackage{xcolor}
\def\BibTeX{{\rm B\kern-.05em{\sc i\kern-.025em b}\kern-.08em
    T\kern-.1667em\lower.7ex\hbox{E}\kern-.125emX}}
    \bibliography{Bibliography.bib}
\begin{document}

\title{Cross Dataset Analysis and Network Architecture Repair for Autonomous Car Lane Detection*\\
\thanks{*We would like to thank Natasha Neogi of NASA LaRC for her feedback on the network architecture repair and cross dataset analysis formulation for this research effort. This work was funded by award 80NSSC20M0005 to the NASA Langley Research Center from the NASA Shared Services Center.}
}

\author{\IEEEauthorblockN{Parth Ganeriwala}
\IEEEauthorblockA{\textit{Department of Computer Science} \\
\textit{Florida Institute of Technology}\\
Melbourne, Florida \\
pganeriwala2022@my.fit.edu}
\and
\IEEEauthorblockN{Siddhartha Bhattacharyya}
\IEEEauthorblockA{\textit{Department of Computer Science} \\
\textit{Florida Institute of Technology}\\
Melbourne, Florida \\
sbhattacharyya@fit.edu}
\and
\IEEEauthorblockN{Raja Muthalagu}
\IEEEauthorblockA{\textit{Department of Computer Science} \\
\textit{BITS Pilani, Dubai Campus}\\
Dubai, United Arab Emirates \\
raja.m@dubai.bits-pilani.ac.in}
% \linebreakand
%  \IEEEauthorblockN{Natasha Neogi}
%  \IEEEauthorblockA{\textit{Langley Research Center} \\
%  \textit{NASA}\\
%  Hampton, Virginia \\
%  natasha.a.neogi@nasa.gov}
}

\maketitle

\begin{abstract}
Transfer Learning has become one of the standard methods to solve problems to overcome the isolated learning paradigm by utilizing knowledge acquired for one task to solve another related one. However, research needs to be done, to identify the initial steps before inducing transfer learning to applications for further verification and explainablity. In this research, we have performed cross dataset analysis and network architecture repair for the lane detection application in autonomous vehicles. Lane detection is an important aspect of autonomous vehicles' driving assistance system. In most circumstances, modern deep-learning-based lane recognition systems are successful, but they struggle with lanes with complex topologies. The proposed architecture, ERF-CondLaneNet is an enhancement to the CondlaneNet used for lane identification framework to solve the difficulty of detecting lane lines with complex topologies like dense, curved and fork lines. The newly proposed technique was tested on two common lane detecting benchmarks, CULane and CurveLanes respectively, and two different backbones, ResNet and ERFNet. 
% The researched technique exhibited comparable performance on both benchmark datasets, while using 33\% less features. 
The researched technique with ERF-CondLaneNet, exhibited similar performance in comparison to Resnet-CondLaneNet, while using 33\% less features, resulting in a reduction of model size by 46\%.
\end{abstract}


\section{Introduction}
Transfer Learning (TL) has been researched extensively in deep learning algorithms for re-purposing models to achieve different tasks without re-training the network repeatedly. A fundamental challenge associated with these supervised deep learning systems are the requirement for large amounts of labeled data, which can be excessively expensive or difficult to obtain in certain cases. Every supervised learning task needs a unique labeled dataset, and training a cutting-edge deep learning model needs substantial computation resources. As a result, Transfer learning has been explored as an option to reduce the training time, improve performance, use less amount of data, thus reducing the computational expense.  
% We define transfer learning as given by Yang et al.\cite{5288526}:

% \begin{quote}
%     Given a source domain $D_S$ and learning task $T_S$, a target domain $D_T$ and learning task $T_T$, transfer learning aims to help improve the learning of the target predictive function $f_T(.)$ in $D_T$ using the knowledge in $D_S$ and $T_S$, where $D_S$ $\neq$ $D_T$, or $T_S$ $\neq$ $T_T$.
% \end{quote}

% In this paper, we propose an inductive deep transfer learning method that can enhance the performance of learning models by incorporating information from a secondary dataset belonging to a similar domain. In the inductive transfer learning setting \cite{5288526}, the target task is different from the source task, no matter whether the source and target domains are the same or not. 
In this work, we introduce ERF-CondLaneNet which is an enhanced lane identification framework with the purpose of solving and easing the difficulty of detecting lane lines with complex topologies such as curved lanes in varying road conditions.  We integrate CondLaneNet \cite{liu_condlanenet_2021} with ERFNet \cite{romera_erfnet_2018} and integrate transfer learning protocols to test on two different lane detecting benchmarks, CULane \cite{CULane_Dataset} and CurveLanes \cite{xu_curvelane-nas_2020} respectively. In this research, we have performed cross-dataset analysis \cite{DBLP:journals/corr/TommasiTC14} to design a method which retains precision while using a significantly less feature space on both benchmark datasets.

% Current research has presented with a variety of lane-detection algorithms with state-of-the-art performance, where some use mathematical models \cite{Bokar11}\cite{wang04} to describe the structure of a given lane whereas others address lane detection as an energy minimization problem \cite{Wojek_Schiele_2008}\cite{hur_multi-lane_2013}.
Current research in developing lane-detection algorithms with state-of-the-art performance has grown exponentially. Some use mathematical models 
% \cite{Bokar11}
\cite{wang04} to describe the structure of a given lane whereas others address lane detection as an energy minimization problem \cite{Wojek_Schiele_2008}.
% \cite{hur_multi-lane_2013}.
Traditional lane identification approaches often use hand-crafted operators to extract features \cite{Tan2014-cm}
% \cite{Jiang2010-xa}\cite{hur_multi-lane_2013} 
and then match the line shape using post-processing techniques such as the Hough transform \cite{Liu2010-oy}\cite{Zhou2010-mk} and Random Sampling Consensus (RANSAC) \cite{Ruyi2011-xt}\cite{4459093}. Others have approached the problem by segmenting the lane using supervised learning models, however, most of these algorithms confine their solutions to recognizing road lanes in a single frame of the driving environment, resulting in poor performance when dealing with demanding driving circumstances such as high shadows, severe road mark deterioration, and extreme vehicle occlusion. In certain cases, the lane may be anticipated in the wrong direction, identified just partially, or not detected at all. One of the key reasons is the information offered by the present frame selected by the researchers is insufficient for accurate lane recognition or prediction. These approaches failed to retain resilience in real-world settings because hand-crafted models cannot cope with the diversity of lane lines in diverse circumstances. 
The objective of our research is as enumerated next:
\renewcommand{\theenumi}{\roman{enumi}}%
\begin{enumerate}
  \item Exploring TL principles with architecture repair while retaining the precision of the existing models. 
  \begin{enumerate}
        \item Extend an existing model with a backbone transformer encoder to decrease the amount of features taken by the supervised learning pipeline.
        \item Maintain a similar or higher F-1 score for benchmark datasets. 
  \end{enumerate}
  \item Unify the machine learning (ML) model by incorporating a diverse range of road conditions.
  \begin{enumerate}
      \item Investigate TL concepts for cross-dataset analysis. 
      \item Perform corresponding repair to the ML network to integrate varying road conditions.
  \end{enumerate}  
  % feature spaces for vehicular systems.
  \item Assess the specifications for architectural change through cross-dataset analysis before applying inductive TL.
\end{enumerate}
% Precisely, the major contribution of this work are:

% \begin{enumerate}
%     \item the enhancement of CondLaneNet with the modification in the backbone encoder as ERFNet, reducing the number of features being used while maintaining the precision of the model; 
%     \item identifying the requirements for the repair of the network after cross-dataset analysis
%     \item implementation of the repair to maintain the F-1 score which involved;
%     \begin{enumerate}
%         \item feature space in cross-domain tasks
%         \item providing similar precision to test sets without having been trained on that dataset's train set;
%     \end{enumerate}
%     \item experimental evaluation of TL concepts for the generation of a unified model that can be tested on untrained datasets;

% \end{enumerate}

% We compare the proposed approach against the CondLaneNet \cite{liu_condlanenet_2021}, and ERFNet \cite{romera_erfnet_2018}. Our experimental evaluation includes two baseline datasets, CULane and CurveLanes, which contains straight lanes and curved lanes on varying road conditions respectively. 
This paper is organised in the following manner. Section II reviews the related works, Section III outlines the proposed methodology and Section IV presents the various benchmark datasets being used and the experimentation conducted. Section V reports the cross dataset analysis and model architecture repair along with the experimental results and discussion. Finally, Section VI concludes our work.

\section{Related Work}

This section describes the current deep-learning-based lane detection systems. Current approaches may be grouped into two groups depending on the strategy of line form description: Convolutional Neural Network (CNN) models, Deep Learning (DL) methods.

\subsection{Convolutional Neural Network Models}
Convolutional neural networks (CNNs) are used in recent lane identification techniques to train deep learning models using popular benchmarks such as TuSimple \cite{noauthor_tusimple_2021} and CULane \cite{CULane_Dataset}. 
Hang et al. \cite{lo_multi-class_2019} proposes two CNN techniques which are Feature Size Selection (FSS) and Degressive Dilation Block (DD Block). They introduced these methodologies to modify the existing semantic segmentation networks. EDANet \cite{lo2019efficient} was chosen as their baseline architecture due to it having a good balance between the efficiency and performance speed for a well defined autonomous driving model. For proper lane localisation, precise geographical information is required and so EDANet features three downsampling processes, whereas most CNNs contain multiple downsampling layers. The modified network achieved an Mean Intersection over Union (MIoU) score of 75.0 on the ITRI dataset.

Liu et al. \cite{liu_lane_2020}, presented a method for increasing the environmental flexibility of the lane detector using Generative Adversarial Networks (GANs) to produce pictures in low-light circumstances. Their suggested approach is divided into three parts: the SIM-CycleGAN, the light conditions style transfer, and the lane identification network. They used ERFNet to evaluate their approaches on the lane detection benchmark CULane and received a 73.9 F-1 score.
Researchers have also worked on formulating feed-forward networks (FFNs) for parameter predictions which are then passed on and trained with a Hungarian fitting loss \cite{liu_end--end_2020}. This end-to-end model outputs parameters of a lane shape model based on a network which is built with a transformer encoder to capture and learn richer features from the images.  
% Spatial CNN (SCNN) is a modified convolutional network proposed by Pan et al.\cite{pan_spatial_2017}, that generalizes typical deep layer-by-layer convolutions to slice-by-slice convolutions inside feature maps, allowing message passings across pixels across rows and columns in a layer. They apply SCNN on Cityscapes dataset with an Intersection over Union (IoU) of 71.6 outperforming Recurrent Neural Network (RNN) based ResNet by 8.7\%. 

% Chng et al.\cite{chng_roneld_2020}, introduces a technique for identifying, tracking, and optimizing active lanes using deep learning probability map outputs using real-time robust neural network output enhancement for active lane identification (RONELD). The network adaptively extracts lane points from probability map outputs, then detects curved and straight lanes before applying weighted least squares linear regression on straight lanes to correct damaged lane edges caused by edge map fragmentation in actual pictures. Finally, they postulate genuine active lanes by tracking previous frames. On cross-dataset validation tests, experimental results show an up to two-fold boost in accuracy when using RONELD.  
According to Wang et al. \cite{wang_multitask_2021}, even if the accuracy of lane line prediction is improving, the capacity of lane markings to localize is rather limited, especially when the lane marking location is remote in nature. They offer a multi-task strategy that combines CNN's network to model semantic information with the high localization ability supplied by handmade features and forecasts the position of the vanishing line. The accuracy of location and network convergence speed are increased by incorporating segmentation, unique handcrafted characteristics, and fitting. Their network outperforms SCNN, ResNet by a huge margin on the benchmark dataset CULane with a F-1 score of 69.6 for Curved Lanes. 
While each of the discussed models performs exceptionally well when trained and tested on the same dataset, its performance suffers dramatically when tested on unknown datasets from various contexts.

\subsection{Deep Learning Methods}

Xu et al. \cite{xu_curvelane-nas_2020}, provide CurveLane-NAS, a novel lane-sensitive architecture search framework for autonomously collecting both long-ranged coherent and accurate short-range curve information. It has three search modules: a feature fusion search module to investigate a better fusion of the local and global context for multi-level hierarchy features; an elastic backbone search module to investigate an efficient feature extractor with good semantics and latency; and an adaptive point blending module to investigate a multi-level post-processing refinement strategy to combine multi-scale head prediction. They also introduce the benchmark called CurveLanes \cite{soulmateb_soulmatebcurvelanes_2021} to include the most problematic curve lanes. It has 150K images and 680K labels and their procedure model achieves an F1-score of 80. 
% Zhang et al.\cite{zhang_end_2018}, presents a deep learning model, Global Convolution Networks (GCN), that handles both classification and localization challenges in semantic lane segmentation. To attain cutting-edge performance, they employ color-based segmentation, a residual-based boundary refinement, and Adam optimization. 
\begin{figure*}[htp]
\centering
\includegraphics[scale=0.5]{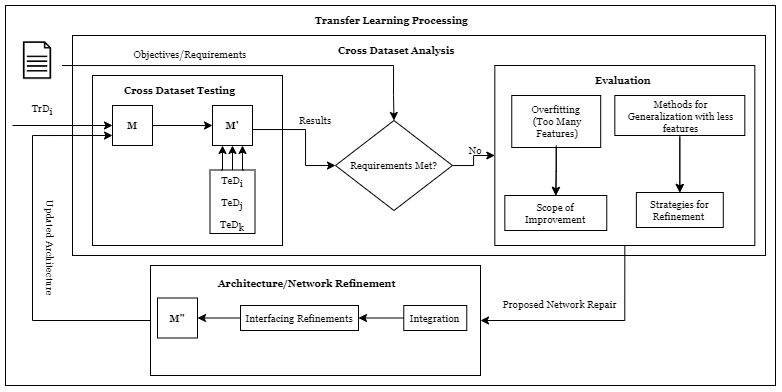}
\caption{Proposed Methodology: CDANR}
\label{fig:method}
\end{figure*}
Researchers \cite{liu_condlanenet_2021} introduce CondLaneNet which is a unique top-to-down lane identification framework that identifies lane instances first and then predicts the line shape for each instance dynamically. The research provides a conditional lane detection technique based on conditional convolution and row-wise formulation to re-solve the lane instance-level discriminating problem. Furthermore, they also introduce the Recurrent Instance Module (RIM) to address the issue of recognizing lane lines with complicated topologies, such as dense lines and fork lines. The advantage from their method is the real-time efficiency and end-to-end pipeline, which requires minimum post-processing. Furthermore, this approach combines accuracy and efficiency, as seen by a 78.14 F1 score and 220 FPS on CULane \cite{CULane_Dataset}. LaneNet is a deep learning module which has been presented by Neven et al. \cite{neven_towards_2018}, which performs end-to-end lane detection by combining binary lane segmentation with a clustering loss function designed for one-shot instance segmentation. The network generates parameters of a perspective transformation where lane fitting is optimal. 
% LaneNet achieves an accuracy of 96.4\% on the TuSimple dataset.      

% Yoo et al.\cite{yoo_end--end_2020}, view the problem not belonging to the instance segmentation field but as finding the set of horizontal locations of each lane marker in the input image. They propose a deep learning model which has been combined with lane marker-wise horizontal reduction modules (HRMs) on top of a end-to-end lane marker detection architecture (E2E-LMD). They achieve an F-1 score of 74.0 against the benchmark CULane and 96.06\% accuracy for TuSimple. Zou et al.\cite{zou_robust_2020}, examine lane detection utilizing numerous frames from a continuous driving scenario and propose a hybrid deep architecture that combines CNN and RNN. A CNN block extracts information from each frame, and the CNN features of numerous continuous frames with the property of time-series are then sent into the RNN block for feature learning and lane prediction.

Wang et al. \cite{wang_lanenet_2018}, devised a lane detecting approach that is comprised of two deep neural networks. The lane edge proposal network uses the initial input image of a vehicle's front view to generate a lane edge proposal map.
The lane line localization network is then in charge of determining the position of each lane given by the lane edge map. 
The use of a deep neural network endows the method with great robustness, and the two-stage detection pipeline reduces computational cost and allows the lane line localization network to be trained in a manner that combines supervised and weakly supervised learning, resulting in a significant reduction in the cost of labeling training data.

\section{Proposed Methodology}

In this section, we elaborate on our proposed methodology Cross Dataset Analysis with Network Refinement (CDANR) shown in Figure \ref{fig:method}, to evaluate a transfer learning approach with architecture/network refinement. 
According to CDANR, initially cross dataset testing needs to be performed on the existing CondLaneNet and ERFNet architecture using the benchmark datasets, CULane and CurveLanes. Then the initial results are evaluated and analyzed to check if it meets the objectives/requirements. 
% In our case, after cross dataset analysis, it was identified that the model was overfitting due to too many features. This led to process of identifying the improvements to the model by integrating strategies to generalise lane topologies while using less number of features. This resulted in the task of network repair. The proposed architectural/network refinement was achieved by enhancing an existing version of CondLaneNet and ERFNet. Changes were made to the backbone of the CondLaneNet model by replacing ResNets with ERFNet \cite{liu_lane_2020}\cite{yoo_end--end_2020}. 
Initial experiments were conducted on both existing architectures ERFNet and CondLaneNet with respect to accuracy of the model as well as the number of features it uses. After evaluating, we found out that while CondLaneNet achieved great accuracy, it was using significantly more features than ERFNet which has been been discussed in further sections. To lay the groundwork for future transfer learning applications, it was necessary to use less parameters, and have a more generic feature extraction which we can obtain from ERFNet when compared to ResNets. Therefore, changes were made to incorporate ERFNet's architecture into the existing architecture as backbone in CondLaneNet.  

\subsection{ERFNet Model}
% Commonly adopted versions of residual blocks as shown in Figure \ref{fig:nonbottleneck} (a) and (b), are employed in a number of recent ConvNets \cite{he_deep_2015} to achieve excellent accuracy in classification and segmentation applications. These residual layer versions have performance problems that can be addressed by the ERFNet model which performs real-time scene segmentation. ERFNet was formulated by sequentially stacking the proposed non-bt-1D layers as shown in Figure \ref{fig:nonbottleneck} (c) that maximizes their learning efficiency and performance. The residual layers utilize factorization \cite{7780677} which is applied to reduce the number of parameters used by the convolutional layers. After applying factorization, the 3x3 convolutional blocks (Figure \ref{fig:nonbottleneck} (a)) are decomposed into equivalent dimensions of each 1D pair giving 3x1 and 1x3 blocks.
% \begin{figure}[h]
% \centering
% \includegraphics[width=\columnwidth]{nonbottleneck.png}
% \caption{\emph{Non-Bottleneck-1d module}}
% \label{fig:nonbottleneck}
% \end{figure}
% \begin{figure*}[h]
% \centering
% \includegraphics[width=\linewidth]{ERFNET_Diagram (4).pdf}
% \caption{\emph{ResNet Model}}
% \label{fig:resnetog}
% \end{figure*}
% \begin{figure*}[h]
% \centering
% \includegraphics[scale=0.45]{ERFNET_Diagram (2).pdf}
% \caption{\emph{Original ERFNet Model}}
% \label{fig:ogerf}
% \end{figure*}
% \begin{figure*}[h]
% \centering
% \includegraphics[width=\linewidth]{ERFNET_Diagram (3).pdf}
% \caption{\emph{Updated ERFNet Model}}
% \label{fig:newerf}
% \end{figure*}
ERFNet is a semantic segmentation architecture which enables the more effective use of parameters, by allowing the network to achieve very high segmentation accuracy even with a reduced feature set while satisfying resource constraints. ERFNet’s implementation of the residual layer, uses the 1D factorization to speed and reduce the parameters of the original non-bottleneck layer, takes advantage of this decomposition. This module is known as ”non-bottleneck-1D” (non-bt-1D), as it is faster and has fewer parameters than the bottleneck design while maintaining the same learning capacity and accuracy. According to Romera et al. \cite{romera_erfnet_2018}, 1D kernels can be factorized for both non-bottleneck and bottleneck implementations. The non-bottleneck architecture clearly benefits more, with a direct 33\% reduction in both convolutions and a significant reduction in execution time. To incorporate it within the CondLaneNet Architecture we made changes to the ERFNet Model. CondLaneNet uses ResNets, as part of their transformer encoder block. The ResNet model, has 34 convolutional layers, with 64 to 2048 feature maps being computed internally through various layers. The original ERFNet model, has 23 convolutional layers with an encoder and decoder block and 8 non-bt-1D convolutional blocks being used. For ResNets, at each residual layer, 256 feature maps are received at the module input, however the non-bt-1D convolutional architecture used in ERFNet, uses only 64 feature maps. This reduces the number of features extracted from the input, while keeping the same performance. The updated ERFNet model was configured to interface with the CondLaneNet architecture and the benchmark datasets. While synchronizing the configurations, 8 additional non-bt-1D bottleneck blocks were added to encase the "extra" features from the input. Also, 2 extra de-convolutional layers were added to upsample the features to the pixel dimensions of the input. Additionally, one downsampling layer and one upsampling layer were added to correlate with the input image dimensions from both the benchmark datasets.

\subsection{CondLaneNet Architecture}

Conditional Lane Detection \cite{liu_condlanenet_2021} is a lane detection method based on conditional convolution, which is type of convolutional configuration containing adjustable kernel parameters that focus on instance-level distinguishing features. This method can be divided into two individual dependent steps: instance detection and shape prediction. 
% \begin{figure*}[h]
% \centering
% \includegraphics[scale=0.5, width=\linewidth]{condlane (1).png}
% \caption{\emph{Enhanced CondLaneNet Architecture}}
% \label{fig:chart3}
% \end{figure*}
For each instance, the instance detection step identifies the object instance and regresses a set of dynamic kernel parameters. Conditional convolutions are used to determine the instance shape in the shape prediction stage. This approach takes advantage of dynamic kernel settings. Shapes may be predicted instance by instance because each instance corresponds to a set of dynamic kernel parameters. This method achieved state-of-the-art performance on instance segmentation tasks \cite{liu_condlanenet_2021}. However, applying the conditional instance segmentation technique directly to lane detection is ineffective. Additionally, due to the extremely large degree of flexibility, segmentation-based shape prediction is inefficient for lane lines \cite{qin_ultra_2020}. 
% On the other hand, the instance detection strategy for general objects is not suitable for slender and curved objects due to the inconspicuous visual characteristic of the border and the center line. 
To address the above-mentioned issues, the conditional lane detection technique enhances shape prediction and instance identification. However, when this architecture was tested with a cross-dataset approach, i.e. trained on $Tr_i$, tested on $Te_j$ and trained on $Tr_j$, tested on $Te_i$, where \emph{i, j} are two different datasets, CondLaneNet gives a poor performance. CULane and CurveLanes were used for the cross-dataset approach and it was observed that both models (one trained on CULane and the other on CurveLanes) perform poorly when tested on CurveLanes and CULane respectively.

\subsection{CondLaneNet Architecture Repair with ERFCondLaneNet}
To the best of our knowledge, through related works, there are generally two strategic machine learning approaches for improving a technique. Either focus is given on improving the accuracy which involves develop complex architectures that are computationally expensive, or the efficiency is improved (reducing model size, feature set decomposition, faster model inference time) by making significant sacrifices in network design in exchange for accuracy. Our approach is focused on improving the core elements of the CondLaneNet architecture: the convolutional blocks. We utilize the conditional lane identification technique, which is based on conditional convolution and row-wise formulation, to address the issue of instance-level discrimination. We investigate re-designing the frequently used residual layers in order to make them more efficient while maintaining equivalent learning performance. Additionally, we evaluate the need for transfer learning as the existing model's accuracy varies when tested on untrained datasets. While this design may be utilized to improve current designs, we present CDANR which is an initial validation step to apply transfer learning to any particular problem. So in essence, we propose initial cross-dataset analysis, followed by architecture change by network repair as a necessity before applying computationally expensive transfer learning techniques.   

\section{Experimentation}
\subsection{Datasets}

To extensively evaluate CDANR, we conducted experiments on three benchmarks: CurveLanes \cite{soulmateb_soulmatebcurvelanes_2021}, CULane \cite{pan_spatial_2017}, and TuSimple \cite{noauthor_tusimple_2021}. CurveLanes is a new benchmark that deals with difficult topologies including fork lines and dense lines. CULane is a big lane detection dataset with nine different scenarios that is regularly utilized. Another extensively used collection of highway driving situations is TuSimple. 
% Table \ref{tab:my-table} displays the details of the three datasets.
% \begin{table}[h]
% \centering
% \begin{tabular}{ccccc}
% \hline
% \textbf{Dataset} & \textbf{Train} & \textbf{Validation} & \textbf{Test} & \textbf{Road Type} \\ \hline 
% CurveLanes & 100K & 20K & 30K & Curved Lane Urban \\ \hline
% CULane & 88.9K & 9.7K & 34.7K & Straight \& Curved Lane Mix  \\ \hline
% TuSimple & 3.3K & 0.4K & 2.8K & Straight Lane Highway \\ \hline\\
% \end{tabular}
% \caption{Datasets for Experimentation}
% \label{tab:my-table}
% \end{table}

\subsubsection{CULane}

CULane is a large-scale, complex dataset for academic traffic lane detection research. It was gathered by cameras set on six separate automobiles throughout Beijing, each driven by a different driver. A total of 133,235 frames were retrieved from more than 55 hours of video. The dataset was split into three sections: 88880 images for training, 9675 images for validation, and 34680 images for testing. The test set was organized into nine categories, each of which corresponds to varying atmospheric and road conditions.
% The traffic lanes were manually labeled by the data collectors using cubic splines for each frame. When lane markers are obscured by cars or are not visible, the lanes were nevertheless labelled based on the situation. The recognition of four lane markings is the emphasis of this dataset, which are the ones that get the most attention in real-world applications.

\subsubsection{TuSimple}

% On the road, there are two categories of items: static objects and dynamic objects. The most static component of the roadway is the lane markers. They show the vehicles how to drive on the road in a pleasant and safe manner. Lane detection is an important feature in autonomous driving since it provides localisation data to the car's control system. This task's dataset comprises video clips with lanes marked on the final frame of each clip. This approach is a time/memory efficient technique that allows better resources utilization for autonomous driving vehicles.

% The dataset contains a variety of complicated properties, including good and moderate weather conditions, various day and night times, and 2-lane/3-lane/4-lane/or more highway highways with varying traffic circumstances.
The dataset was partitioned into training and testing sets, including 3626 video clips and 3626 annotated frames in training and 2782 video clips in testing. The clips are each a one-second clip with 20 frames, with the camera's view direction fairly close to the driving direction. Poly lines for lane markings are the annotations. Although most lanes have four lane markings (current lane, left/right lanes), there are at most five lane markings for some lanes. When changing lanes, the extra lane is employed since it is difficult to discern which lane is the current one.

\subsubsection{CurvedLanes}

CurveLanes is a new benchmark lane detection dataset with 150K lanes pictures with a diverse range of traffic lane detection scenarios like curves and multi-lanes. It was gathered in numerous Chinese cities in real-world urban and highway environments. It's the world's largest lane detection dataset to date, and it sets a higher bar for the machine learning community.

The entire 150K dataset is divided into three sections: train: 100K, validation: 20K, and testing: 30K. The majority of the photos in this dataset have a resolution of 2650 $\times$ 1440 pixels.

\section{Results and Discussion}
\subsection{Validating Existing CondLaneNet and ERFNet}
Cross dataset evaluation is very uncommon due to the compatibility issues between the data. In this work, we performed cross dataset analysis to assess whether architectural changes are a definitive step before inducing transfer learning. To do this, the model with existing CondLaneNet architecture was trained on the CurveLanes Dataset and then tested with the CULane dataset. Consequently, the model was trained on CULane dataset and tested on CurveLanes. A similar process flow was taken for the TuSimple dataset as well. The training testing ratio was taken as 70:30 and uniform test sets have been used in this analysis.
% ERFNET and RESNET Difference. 

We also trained the ERFNet model using the CULane Dataset, and after testing, we obtained a 97.78\% IoU (Intersection over Union) as shown in Table \ref{tab:init1}. A total F-1 score of 0.7357 was recorded for CULane. Among the 9 categories present in the CULane Dataset, for normal road conditions was 0.9170 and for curved road conditions it was 0.6672.

For TuSimple Dataset which was trained and tested with ERFNet, we recorded a 98.35\% IoU and the accuracy was 93.39\%. However, due to TuSimple not having an extensive amount of features in comparison to CULane, it was disregarded and not used in further evaluation. 
\begin{table}[h]
\centering
\begin{center}
\caption{\emph{Initial Analysis by Cross Dataset Evaluation}}
 \begin{tabular}{c c c} 
 \hline
 \multicolumn{3}{ c }{ERFNet Architecture} \\
 \hline
 Training Dataset & Testing Dataset & IoU  \\ [1ex] 
 \hline\hline
 CurveLanes & CurveLanes & 98.59\% \\ 
 \hline
  CurveLanes & CULane & 96.34\% \\ 
 \hline
  CULane & CULane & 97.78\% \\ 
 \hline
  CULane & CurveLanes & 72.45\% \\ 
 \hline\linebreak
\end{tabular}
\label{tab:init1}
\end{center}

\end{table}

\subsection{Initial Cross Dataset Analysis}
For the initial analysis as shown in Table \ref{tab:init}, we see that the CondLaneNet Architecture Model trained on CULane gives a moderately good F-1 Score of 0.7948 when tested on itself. It is also observed that the model when trained on CurveLanes gives a good F-1 Score of 0.8610 when tested on itself. In the first case, when the model was trained on CurveLanes and tested on CULane, we found that the F-1 score was 0.6545 but upon further investigation the recall and precision were balanced.  Thus, we assume that there was a training issue and it requires different types of data to be included. For the second case where it was trained on CULane and tested on CurveLanes, we see an unusual performance where the F-1 score was 0.526 but the precision was 0.8168 and recall was 0.388 and an imbalance in recall and precision was noted. This explains that the model was not able to predict/detect curved lanes in entirety but the ones it does detect are accurate given by the high precision. In our proposed approach, we investigated architectural changes with transfer learning, if the problem can be solved. 
% This can be addressed with the architectural change of introducing ERFNets as a backbone instead of ResNets.   
\begin{table}[h]
\centering
\begin{center}
\caption{\emph{Initial Analysis by Cross Dataset Evaluation}}
 \begin{tabular}{c c c} 
 \hline
 \multicolumn{3}{ c }{CondLaneNet Architecture} \\
 \hline
 Training Dataset & Testing Dataset & F-1 Score  \\ [1ex] 
 \hline\hline
 CurveLanes & CurveLanes & 0.8610 \\ 
 \hline
  CurveLanes & CULane & 0.6545 \\ 
 \hline
  CULane & CULane & 0.7948 \\ 
 \hline
  CULane & CurveLanes & 0.5260 \\ 
 \hline\linebreak
\end{tabular}
\label{tab:init}
\end{center}

\end{table}

\subsection{Architecture Repair \& Cross Dataset Analysis}
Following, the introduction of our proposed architecture, ERFNet retained the F-1 score in other scenarios as shown in the Table \ref{tab:final}. ERFNet, due to the improved bottleneck method takes less amount of parameters during training when compared to ResNets. With 33\% less feature parameters being used, it resulted in significantly reduced while maintaining the F-1 score measure. Table \ref{tab:size} shows the model size recorded before and after the architectural changes with a reduced model size by 50\%. 
\begin{table}[h]
\centering
\begin{center}
\caption{\emph{Analysis by Cross Dataset Evaluation after Architectural Change}}
 \begin{tabular}{c c c} 
 \hline
 \multicolumn{3}{ c }{ERF-CondLaneNet Architecture} \\
 \hline
 Training Dataset & Testing Dataset & F-1 Score  \\ [1ex] 
 \hline\hline
 CurveLanes & CurveLanes & 0.8467 \\ 
 \hline
  CurveLanes & CULane & 0.6915 \\ 
 \hline
  CULane & CULane & 0.7513 \\ 
 \hline
  CULane & CurveLanes & 0.3320 \\ 
 \hline\linebreak
\end{tabular}
\label{tab:final}
\end{center}

\end{table}
We see a slight increase in F-1 score to 0.6915, when the model was trained on CurveLanes and tested on CULane. This is due to the fact that straight lines are a subset of curved lanes.
% domain encompassing the features required to predict straight lane lines. 
There was a further decrease in F-1 score associated with the CULane vs CurveLanes test. We assume that models trained on straight lanes cannot detect and predict curved lanes as can be seen with decreased performance for both ResNets and ERFNet. The exponential decrease in the F-1 score for ERFNet is assumed to be due to usage of lesser features based on the bottleneck method which led to lesser F-1 Score in the last case. It is expected to maintain the accuracy, an inductive transfer learning architecture will be required to align the knowledge with the learning system. 
% Yang et al. \cite{5288526} defined the inductive transfer learning setting where the target task is different from the source task, but the source and target domains remain the same. 

\begin{table}[h]
\centering
\begin{center}
\caption{\emph{Model Size before and after Architectural Change}}
 \begin{tabular}{c c c} 
 \hline
  \multicolumn{3}{ c }{Backbone Used in Architecture} \\
 \hline
 Model & ERFNet & ResNet  \\ [1ex] 
 \hline\hline
 Size after training for 14 epochs & 563 MB & 1.2 GB \\
\end{tabular}
\label{tab:size}
\end{center}

\end{table}
% REVISIT THIS
We can see, that using less amount of parameters that is, extracting features from a more general aspect rather than diving into specific features gives a better understanding for transfer learning between different systems. 

% After the architectural change was made, the findings demonstrate that our technique can handle complicated line topologies. Even for dense lines and fork lines, our technique can successfully distinguish the occurrences. We can further explore inductive transfer learning to scale the accuracy to other topologies and varying road conditions. 

% CROSS dataset analysis, 
\section{Conclusion}
In this study, a new architecture was proposed that consisted of ERFNet-CondLaneNet, which is an integrated car lane detection architecture with ERFNet's semantic segmentation network as the backbone. Cross dataset testing was performed to evaluate if existing architectures met the objectives. The initial results demonstrated poor performance while being computationally expensive by the existing architectures. After evaluation, refinements were proposed to the network architecture, that was then integrated and interfaced with the architecture. Using the proposed architecture, cross dataset analysis was carried out on benchmark datasets CULane and CurveLanes. During the implementation of the model, it was found out the new architecture uses less amount of parameters giving a significantly smaller model size, while retaining the same accuracy as other models in the car lane detection domain. Therefore, with a generalised feature extraction, where specific features are not considered, we can indeed retain the same accuracy. This lays the groundwork to further transfer learning applications for cross-vehicular lane detection systems. Although, cross-vehicular lane detection systems perform the same task, the semantic of lanes changes. For example, automobiles have lanes on the road which they have to stay within, aircrafts have lanes to stay within and the central line to adhere to.
% \cite{meymandinejad}.

% This gives way to the future transfer learning applications, where we make changes to the last few layers of a model, leading to small architectural changes as well as different weight instantiation and initialization. With a new problem, we can simply utilize the off-the-shelf features of a CNN pre-trained on various models and train a new model on these extracted features. In practice, pre-trained features are typically employed for adaption scenarios, when we wish to adjust to a new task.

% With the increase of the autonomous vehicular market, governments throughout the world have made features like automated emergency braking (AEB) and lane departure warning (LDW) mandatory, making way for advanced technologies to be developed and deployed. Many safety measures have been developed to assist drivers and limit the amount of accidents. Lane detection, in particular is safety critical for all of these features to align with each other and work in collaboration. To this end, autonomous cars should be trained on a diverse range of datasets from different regions for accurate predictions instead of training them on roadways from just a particular region. As autonomous systems such as Tesla become more mainstream, testing vision data disclosure becomes important for safety assurance and verification. Our cross dataset analysis technique provides a way to validate how well the model generalizes on unknown scenarios which it has not been trained upon.

\printbibliography

@inproceedings{Zhou2010-mk, title           = "A novel lane detection based on geometrical model and
                     Gabor filter",
  booktitle       = "2010 {IEEE} Intelligent Vehicles Symposium",
  author          = "Zhou, Shengyan and Jiang, Yanhua and Xi, Junqiang and
                     Gong, Jianwei and Xiong, Guangming and Chen, Huiyan",
  publisher       = "IEEE", month           =  jun,
  year            =  2010,
  conference      = "2010 IEEE Intelligent Vehicles Symposium (IV)",
  location        = "La Jolla, CA, USA"
}

@article{4459093,  author={Kim, ZuWhan},  journal={IEEE Transactions on Intelligent Transportation Systems},   title={Robust Lane Detection and Tracking in Challenging Scenarios},   year={2008},  volume={9},  number={1},  pages={16-26},  doi={10.1109/TITS.2007.908582}}

@misc{lo2019efficient,
      title={Efficient Dense Modules of Asymmetric Convolution for Real-Time Semantic Segmentation}, 
      author={Shao-Yuan Lo and Hsueh-Ming Hang and Sheng-Wei Chan and Jing-Jhih Lin},
      year={2019},
      eprint={1809.06323},
      archivePrefix={arXiv},
      primaryClass={cs.CV}
}

@article{Ruyi2011-xt,
  title     = "Lane detection and tracking using a new lane model and distance
               transform",
  author    = "Ruyi, Jiang and Reinhard, Klette and Tobi, Vaudrey and Shigang,
               Wang",
  journal   = "Mach. Vis. Appl.",
  publisher = "Springer Science and Business Media LLC",
  volume    =  22,
  number    =  4,
  pages     = "721--737",
  month     =  jul,
  year      =  2011,
  language  = "en"
}

@inproceedings{Tan2014-cm,
  title           = "A novel curve lane detection based on Improved River Flow
                     and {RANSA}",
  booktitle       = "17th International {IEEE} Conference on Intelligent
                     Transportation Systems ({ITSC})",
  author          = "Tan, Huachun and Zhou, Yang and Zhu, Yong and Yao, Danya
                     and Li, Keqiang",
  publisher       = "IEEE",
  month           =  oct,
  year            =  2014,
  conference      = "2014 IEEE 17th International Conference on Intelligent
                     Transportation Systems (ITSC)",
  location        = "Qingdao, China"
}

@inproceedings{Liu2010-oy,
  title           = "Combining Statistical Hough Transform and Particle Filter
                     for robust lane detection and tracking",
  booktitle       = "2010 {IEEE} Intelligent Vehicles Symposium",
  author          = "Liu, Guoliang and Worgotter, Florentin and Markelic, Irene",
  publisher       = "IEEE",
  month           =  jun,
  year            =  2010,
  conference      = "2010 IEEE Intelligent Vehicles Symposium (IV)",
  location        = "La Jolla, CA, USA"
}

@article{wang_multitask_2021,
	title = {Multitask Attention Network for Lane Detection and Fitting},
	issn = {2162-237X, 2162-2388},
	url = {https://ieeexplore.ieee.org/document/9286879/},
	doi = {10.1109/TNNLS.2020.3039675},
	pages = {1--13},
	journaltitle = {{IEEE} Transactions on Neural Networks and Learning Systems},
	shortjournal = {{IEEE} Trans. Neural Netw. Learning Syst.},
	author = {Wang, Qi and Han, Tao and Qin, Zequn and Gao, Junyu and Li, Xuelong},
	urldate = {2021-10-08},
	date = {2021},
	langid = {english},
}

@article{liu_condlanenet_2021,
	title = {{CondLaneNet}: a Top-to-down Lane Detection Framework Based on Conditional Convolution},
	url = {http://arxiv.org/abs/2105.05003},
	shorttitle = {{CondLaneNet}},
	journaltitle = {{arXiv}:2105.05003 [cs]},
	author = {Liu, Lizhe and Chen, Xiaohao and Zhu, Siyu and Tan, Ping},
	urldate = {2021-10-08},
	date = {2021-06-10},
	eprinttype = {arxiv},
	eprint = {2105.05003},
}

@article{neven_towards_2018,
	title = {Towards End-to-End Lane Detection: an Instance Segmentation Approach},
	url = {http://arxiv.org/abs/1802.05591},
	shorttitle = {Towards End-to-End Lane Detection},
	journaltitle = {{arXiv}:1802.05591 [cs]},
	author = {Neven, Davy and De Brabandere, Bert and Georgoulis, Stamatios and Proesmans, Marc and Van Gool, Luc},
	urldate = {2021-10-08},
	date = {2018-02-15},
	eprinttype = {arxiv},
	eprint = {1802.05591},
}

@article{liu_end--end_2020,
	title = {End-to-end Lane Shape Prediction with Transformers},
	url = {http://arxiv.org/abs/2011.04233},
	journaltitle = {{arXiv}:2011.04233 [cs]},
	author = {Liu, Ruijin and Yuan, Zejian and Liu, Tie and Xiong, Zhiliang},
	urldate = {2021-10-08},
	date = {2020-11-28},
	eprinttype = {arxiv},
	eprint = {2011.04233},
}

@article{wang_lanenet_2018,
	title = {{LaneNet}: Real-Time Lane Detection Networks for Autonomous Driving},
	url = {http://arxiv.org/abs/1807.01726},
	shorttitle = {{LaneNet}},
	journaltitle = {{arXiv}:1807.01726 [cs]},
	author = {Wang, Ze and Ren, Weiqiang and Qiu, Qiang},
	urldate = {2021-10-08},
	date = {2018-07-04},
	eprinttype = {arxiv},
	eprint = {1807.01726},
}

@article{qin_ultra_2020,
	title = {Ultra Fast Structure-aware Deep Lane Detection},
	url = {http://arxiv.org/abs/2004.11757},
	journaltitle = {{arXiv}:2004.11757 [cs]},
	author = {Qin, Zequn and Wang, Huanyu and Li, Xi},
	urldate = {2021-12-16},
	date = {2020-08-04},
	eprinttype = {arxiv},
	eprint = {2004.11757},
}

@inproceedings{Wojek_Schiele_2008, place={Berlin, Heidelberg}, series={Lecture Notes in Computer Science}, title={A Dynamic Conditional Random Field Model for Joint Labeling of Object and Scene Classes}, ISBN={9783540886938}, DOI={10.1007/978-3-540-88693-8_54}, abstractNote={Object detection and pixel-wise scene labeling have both been active research areas in recent years and impressive results have been reported for both tasks separately. The integration of these different types of approaches should boost performance for both tasks as object detection can profit from powerful scene labeling and also pixel-wise scene labeling can profit from powerful object detection. Consequently, first approaches have been proposed that aim to integrate both object detection and scene labeling in one framework. This paper proposes a novel approach based on conditional random field (CRF) models that extends existing work by 1) formulating the integration as a joint labeling problem of object and scene classes and 2) by systematically integrating dynamic information for the object detection task as well as for the scene labeling task. As a result, the approach is applicable to highly dynamic scenes including both fast camera and object movements. Experiments show the applicability of the novel approach to challenging real-world video sequences and systematically analyze the contribution of different system components to the overall performance.}, booktitle={Computer Vision – ECCV 2008}, publisher={Springer}, author={Wojek, Christian and Schiele, Bernt}, editor={Forsyth, David and Torr, Philip and Zisserman, Andrew}, year={2008}, pages={733–747}, collection={Lecture Notes in Computer Science} }

@article{wang04,
author = {Wang, Yue and Teoh, Eam and Shen, Dinggang},
year = {2004},
month = {04},
pages = {269-280},
title = {Lane detection and tracking using B-Snake},
volume = {22},
journal = {Image and Vision Computing},
doi = {10.1016/j.imavis.2003.10.003}
}

@article{pan_spatial_2017,
	title = {Spatial As Deep: Spatial {CNN} for Traffic Scene Understanding},
	url = {http://arxiv.org/abs/1712.06080},
	shorttitle = {Spatial As Deep},
	journaltitle = {{arXiv}:1712.06080 [cs]},
	author = {Pan, Xingang and Shi, Jianping and Luo, Ping and Wang, Xiaogang and Tang, Xiaoou},
	urldate = {2021-10-08},
	date = {2017-12-17},
	eprinttype = {arxiv},
	eprint = {1712.06080},
}

@article{romera_erfnet_2018,
	title = {{ERFNet}: Efficient Residual Factorized {ConvNet} for Real-Time Semantic Segmentation},
	volume = {19},
	issn = {1524-9050, 1558-0016},
	url = {http://ieeexplore.ieee.org/document/8063438/},
	doi = {10.1109/TITS.2017.2750080},
	shorttitle = {{ERFNet}},
	pages = {263--272},
	number = {1},
	journaltitle = {{IEEE} Transactions on Intelligent Transportation Systems},
	shortjournal = {{IEEE} Trans. Intell. Transport. Syst.},
	author = {Romera, Eduardo and Alvarez, Jose M. and Bergasa, Luis M. and Arroyo, Roberto},
	urldate = {2021-10-08},
	date = {2018-01},
	langid = {english},
}

@article{xu_curvelane-nas_2020,
	title = {{CurveLane}-{NAS}: Unifying Lane-Sensitive Architecture Search and Adaptive Point Blending},
	url = {http://arxiv.org/abs/2007.12147},
	shorttitle = {{CurveLane}-{NAS}},
	journaltitle = {{arXiv}:2007.12147 [cs]},
	author = {Xu, Hang and Wang, Shaoju and Cai, Xinyue and Zhang, Wei and Liang, Xiaodan and Li, Zhenguo},
	urldate = {2021-10-08},
	date = {2020-07-23},
	eprinttype = {arxiv},
	eprint = {2007.12147},
}

@article{liu_lane_2020,
	title = {Lane Detection in Low-light Conditions Using an Efficient Data Enhancement : Light Conditions Style Transfer},
	url = {http://arxiv.org/abs/2002.01177},
	shorttitle = {Lane Detection in Low-light Conditions Using an Efficient Data Enhancement},
	journaltitle = {{arXiv}:2002.01177 [cs]},
	author = {Liu, Tong and Chen, Zhaowei and Yang, Yi and Wu, Zehao and Li, Haowei},
	urldate = {2021-10-08},
	date = {2020-05-16},
	eprinttype = {arxiv},
	eprint = {2002.01177},
}

@article{lo_multi-class_2019,
	title = {Multi-Class Lane Semantic Segmentation using Efficient Convolutional Networks},
	url = {http://arxiv.org/abs/1907.09438},
	journaltitle = {{arXiv}:1907.09438 [cs]},
	author = {Lo, Shao-Yuan and Hang, Hsueh-Ming and Chan, Sheng-Wei and Lin, Jing-Jhih},
	urldate = {2021-10-08},
	date = {2019-07-22},
	eprinttype = {arxiv},
	eprint = {1907.09438},
}

@misc{CULane_Dataset, title = {CuLane Dataset}, url={https://xingangpan.github.io/projects/CULane.html}, }

@software{noauthor_tusimple_2021,
	title = {{TuSimple} Competitions for {CVPR}2017},
	rights = {Apache-2.0},
	url = {https://github.com/TuSimple/tusimple-benchmark},
	publisher = {{TuSimple}},
	urldate = {2021-10-08},
	date = {2021-10-08},
	note = {original-date: 2017-05-06T21:18:47Z},
}

@software{soulmateb_soulmatebcurvelanes_2021,
	title = {{SoulmateB}/{CurveLanes}},
	url = {https://github.com/SoulmateB/CurveLanes},
	author = {{SoulmateB}},
	urldate = {2021-10-08},
	date = {2021-10-07},
	note = {original-date: 2020-07-28T01:58:05Z},
}

@article{DBLP:journals/corr/TommasiTC14,
  author    = {Tatiana Tommasi and
               Tinne Tuytelaars and
               Barbara Caputo},
  title     = {A Testbed for Cross-Dataset Analysis},
  journal   = {CoRR},
  volume    = {abs/1402.5923},
  year      = {2014},
  url       = {http://arxiv.org/abs/1402.5923},
  eprinttype = {arXiv},
  eprint    = {1402.5923},
  bibsource = {dblp computer science bibliography, https://dblp.org}
}

\end{document}